%% file: main.tex
\documentclass{wscpaperproc}
\usepackage{latexsym}
\usepackage{caption}
\usepackage{graphicx}
\usepackage{mathptmx}
\usepackage[T1]{fontenc}
\usepackage{tabularray}
\usepackage{subcaption}
\usepackage[autostyle]{csquotes} 
\usepackage[pdftex,colorlinks=true,urlcolor=blue,citecolor=black,
anchorcolor=black,linkcolor=black]{hyperref}
\usepackage{amsmath}
\usepackage{amsfonts}
\usepackage{amssymb}
\usepackage{amsbsy}
\usepackage{amsthm}

\usepackage[pdftex,colorlinks=true,urlcolor=blue,citecolor=black,anchorcolor=black,linkcolor=black]{hyperref}

\newtheoremstyle{wsc}
{3pt}
{3pt}
{}
{}
{\bf}
{}
{.5em}
{}

\theoremstyle{wsc}

\begin{document}

%
%

\pagestyle{fancyplain}

\thispagestyle{plain}
\firstPageHead{}

\chead{\fancyplain{}{\itshape Kuiper, Lin, Blanchet, and Tarokh}}

\rhead{}
\cfoot{}
\renewcommand{\headrulewidth}{0pt} 

\input{wscbib.tex}           

\setlength{\baselineskip}{12.7pt}

\title{Generative Learning for Simulation of Vehicle Faults}

\author{\begin{center}Patrick Kuiper\textsuperscript{1}, Sirui Lin\textsuperscript{2}, Jose Blanchet\textsuperscript{2}, and Vahid Tarokh\textsuperscript{1}\\
[11pt]
\textsuperscript{1}Dept.~of Electrical and Computer Eng., Duke University, Durham, NC, USA\\
\textsuperscript{2}Dept.~of Management Science and Eng., Stanford University, Stanford, CA, USA\end{center}
}

\maketitle

\vspace{-12pt}

\section*{ABSTRACT}
We develop a novel generative model to simulate vehicle health and forecast faults, conditioned on practical operational considerations. The model, trained on data from the US Army's Predictive Logistics program, aims to support predictive maintenance. It forecasts faults far enough in advance to execute a maintenance intervention before a breakdown occurs. The model incorporates real-world factors that affect vehicle health. It also allows us to understand the vehicle's condition by analyzing operating data, and characterizing each vehicle into discrete states. Importantly, the model predicts the time to first fault with high accuracy. We compare its performance to other models and demonstrate its successful training. 

\section{INTRODUCTION}
\label{sec:intro}

Many organizations have realized the utility of using machine learning and a data driven approach to understand and improve the health of their vehicle fleets; these organizations invest enormous amounts of resources in transportation systems as they are critical to operations. We focus this analysis on the United States' Department of Defense (DoD), where the US Army alone is projected to spend an estimated \$5 billion per year (in 2020 dollar terms through 2050),  developing and acquiring ground vehicles, where ground vehicles are any vehicles other than aircraft and ships \cite{cbo_ground_sys}. Maintaining this enormous investment is critical to ensuring combat readiness across the DoD, where the department spent \$90 billion in 2022 on maintaining vehicles across domains: ground, air, and sea \cite{gao_pred_maint}. 

Predicting requirements is critical to an effective maintenance program. The application of statistics towards vehicle maintenance prediction is often referred to as predictive maintenance. Recognizing the importance of predictive maintenance, in the 2022 National Defense Authorization Act (NDAA) Congress required the DoD Inspector General Office to review predictive maintenance practices, originally established by DoD directives in 2002 and 2007 \cite{dod_ig}. Given the importance of predictive maintenance in promoting readiness and conserving budgetary resources, the US Army is employing hardware and software solutions to address Congressional and Departmental guidance.

 We propose a model designed to achieve \emph{predictive maintenance}, with the goal of completing corrective actions on a vehicle in order to avoid more costly or catastrophic damage in the future \cite{pred_maint_survey}. This predictive maintenance model is \emph{data driven}, where conclusions are based on the outcomes observed in a dataset of sensor and fault signals recorded from an onboard vehicle computer. This data was collected by the US Army under an effort designated as Condition Based Maintenance (CBM). In this effort millions of time indexed data points were collected from a number of military vehicle types, with the goal of informing the development of maintenance processes, hardware, and analysis software to support predictive maintenance. We have developed a generative learning approach trained on a publicly releasable subset of this CBM data to predict future vehicle faults. Additionally, we simulate outcomes conditioned on practical vehicle considerations, including vehicle age, location, and mode sub-type.

We begin the discussion of our model with a brief literature review, introducing general time-series modeling, then focusing on prediction models for vehicle faults, and relevant technical reports related to the maintenance of military vehicles. Next, we review an analysis related to the DoD's efforts with predictive maintenance and the CBM dataset used to develop our model. We then introduce the proposed generative fault prediction model. We conclude with results, providing prediction performance analysis, comparison to alternate models, and interpretation of the analysis provided by the proposed model. Additional information regarding model training convergence and parameter interpretation are provided in the appendices, including a link to a github repository with all modeling code and processor information in Appendix \ref{app:code}.  In summary, our contributions include:

\begin{enumerate}
\item We develop and train a practical, generative learning model to accurately predict vehicle faults and time to first fault on a rolling basis in real time.
\item We generate future sensor feature covariates based on the values of previous covariates, using an attention based transformer.
\item We also generate future sensor feature covariates based on the values of previous data and a selected vehicle use profile, where the profile is based upon practical considerations, using a Variational Auto-Encoder and K-means clustering.

\item We provide results and conclusions which are interpretable and understandable among professional practitioners in the application field.
\end{enumerate}

\section{Related Work}

Time series analysis, which leverages time-indexed data to extract meaningful statistical conclusions, remains an important field in statistics with applications in finance, medicine, and manufacturing. The wide adoption of machine learning techniques, specifically generative learning, has led to the development of a number of innovative time series methods. We discuss these techniques, along with their application to reliability engineering in automobiles.

\subsection{General Time Series Prediction}
\label{sec:ts_pred}
A number of effective, well established methods exist to perform time series analysis; these include Auto-Regressive (AR) and Moving Average (MA) models, in addition to more advanced integration of both AR and MA, with Auto-Regressive Moving Average (ARMA) and Auto-Regressive Integrated Moving Average (ARIMA) \cite{boxjen76}.  Approaches employing neural networks specifically pertaining to the forecasting of time-indexed sequence data have included Recurrent Neural Networks (RNNs) and Long-Short Term Memory networks (LSTMs). Of particular interest to this analysis is the Deep-Auto Regressive (DeepAR) method developed by \citeN{salinas2019deepar}. The DeepAR method employs a LSTM-based neural network design, with an AR component, that learns from multiple time series data points. In the proposed model, multiple time series data includes the fault target and sensor feature covariates. 

The DeepAR method produces a probabilistic function via Monte Carlo simulation, allowing for the model to be queried on quantiles, which then generates forecasted future data. This enables the generation of multiple future predictions via the quantile function, improving the utility of the analysis when compared to point estimates. A critical consideration is that the DeepAR model requires the incorporation of \emph{future} covariates when forecasting a future target variable. A novel approach for generating multivariate time series predictions is proposed by \shortciteN{attn_mech}, with the use of a spatio-temporal attention mechanism (STAM).  This model isolates critical time steps (temporal) and variables (spatial) when forecasting future target predictions. It does not depend on future covariates during the prediction step as with DeepAR; however, we found STAM alone does not perform as well on forecasting vehicle faults when compared to the DeepAR model (see Section \ref{sec:results} for comparison). We propose a hybrid model using both STAM and DeepAR, where the requirement of the DeepAR model to incorporate future covariates is addressed by generating these sensor values using STAM. We develop this approach further in the Section \ref{sec:cond_model} with additional discussion in Section \ref{sec:compare}. 

\subsection{Vehicle Fault Prediction}

Using generative learning to anticipate faults in vehicles remains a relatively new topic. Recently, \citeN{mult_time_series} completed an analysis similar to our proposed application, using self-supervised learning and a graph based technique to forecast vehicle faults. The formulation of the technique developed by \citeN{mult_time_series} differ significantly to our proposed method, and the dataset is much smaller, with the faults being labeled by experts visually reviewing sensor signals. This is in contrast to our analysis which leverages a much larger dataset collected in non-experimental, field conditions, where the faults are labeled by an automated vehicle onboard computer. 

Another similar analysis is provided by \citeN{fault_class_discrim}, where several discriminative classification methods are evaluated for determining the occurrence of vehicle faults in several different subsystems. The evaluated classification model takes real-time vehicle data in the form of sensor data and Diagnostic Trouble Codes (DTC), with dimensional reduction using Principal Component Analysis (PCA). The experiments proposed in this analysis are similar to our proposal in that data was collected from dozens of vehicles of a single vehicle family (make / model), with the goal of forecasting faults in real time using machine learning. However, the model only produces a discriminative classification, not a generative prediction, and no interpretable information is provided about the causes of vehicle failure from this analysis.

A general review of machine learning applications for vehicle predictive maintenance is proposed by \citeN{pred_maint_survey}. Of note from this rigorous review and categorization of related work is the proposal of several research challenges in the field. The authors highlight the need for \emph{public real-world datasets}, which are \emph{labeled}. An additional relevant challenge referenced in this work is the use of more complex, \emph{truly deep neural networks} for modeling, while maintaining \emph{interpretability} and \emph{explainability} given the need for understanding when addressing maintenance concerns. We work to address all of these research challenges in this proposal and provide a practical model for a large family of vehicles in the US Army, which may be readily applied to additional vehicle types.

\subsection{Military Maintenance Background}

An detailed report of the CBM dataset, along with a number of other predictive maintenance efforts established by the US Navy and Marine Corps, is provided by \citeN{rit_report}. This report provides a wealth of technical detail concerning the vehicle population covered in the CBM dataset, along with the entire DoD predictive maintenance effort across the uniformed services. 

As a check on the CBM dataset, we reference \citeN{orcen_report} who developed an insightful analysis into historical operational readiness rates of several common US Army vehicles. This analysis is relevant as it describes with precision common operational readiness rates observed among organizations in the US Army. These published readiness rates serve as a validation for observed rates recorded in the CBM dataset used in the proposed analysis.

\section{CBM Dataset Overview}

We have identified a research demand for a real-world dataset composed of vehicle sensor readings, with associated vehicle faults. Such a dataset would allow for the development of a data driven fault prediction model. This dataset would be of particular use in military and civilian automotive applications. We now provide further details about the US Army CBM dataset which we propose meets the challenges presented by the predictive maintenance research community, and further develop our assumptions that vehicle electronic fault signals may be used as a proxy for vehicle condition.  

\subsection{US Military Predictive Maintenance Data Collection Efforts}

According to the Government Accountability Office (GAO) report titled ``Military Readiness: Actions Needed to Further Predictive Maintenance on Weapon Sytems'', published in December 2022 \cite{gao_pred_maint}, the US DOD's predictive maintenance efforts may be summarized as
\blockquote{...any effort that uses condition-monitoring technology or analysis of historical data to anticipate maintenance needs in a manner that reduces unscheduled reactive maintenance or overly prescriptive preventive maintenance. The military services use multiple terms for predictive maintenance or predictive maintenance enablers, including Condition-Based Maintenance (CBM), Condition-Based Maintenance Plus (CBM+), Prognostic and Predictive Maintenance (PPMx), Enhanced Reliability-Centered Maintenance (eRCM), and Predictive Maintenance.}

\begin{figure}[h!]
    \centering
    \includegraphics[width=8cm]{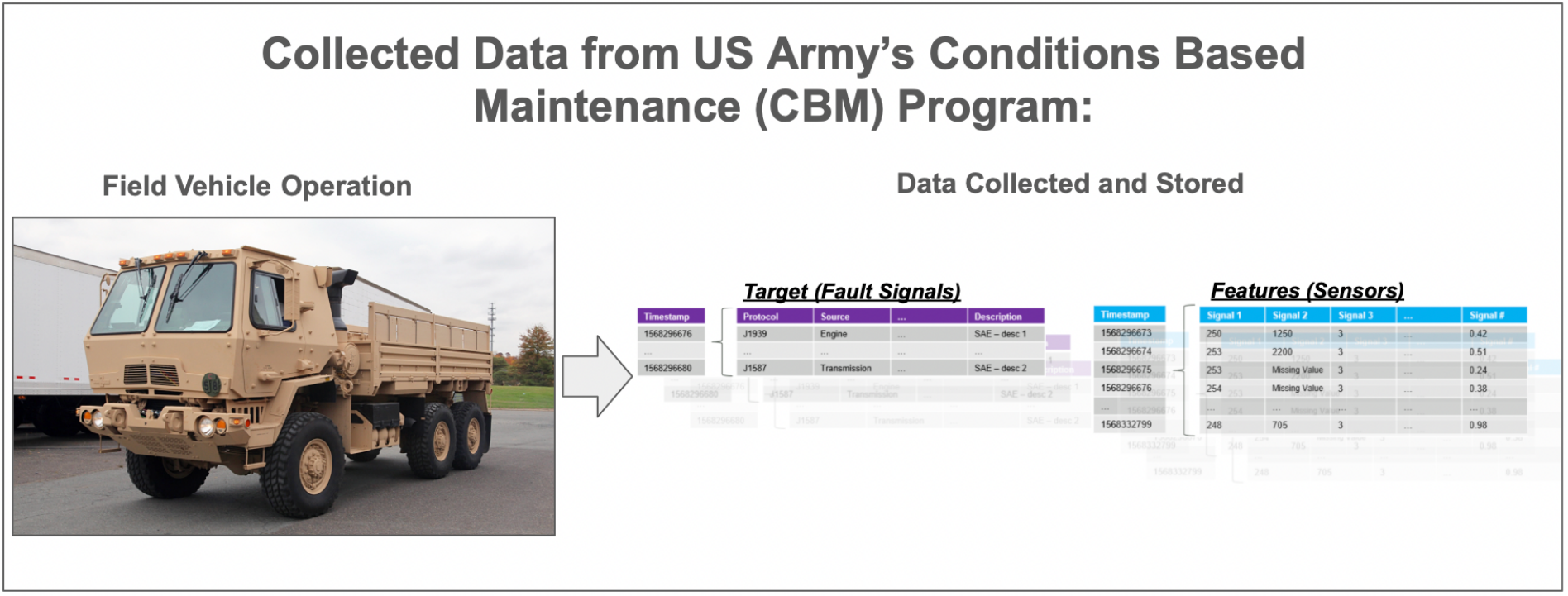}
    \caption{Description of Condition Based Maintenance dataset. Data from field vehicle operations, collected over several months from various vehicle types, are presented. This dataset includes time series of various fault signals, each with associated recording timestamps, as well as time series from multiple sensors also timestamped. Our analysis aims to utilize the sensor data as feature covariates to predict fault occurrences.}
    \label{fig:enter-label1}
\end{figure}

The dataset used in this analysis is from the US Army's CBM effort. As illustrated in Figure~\ref{fig:enter-label1}, sensor and fault data was recorded over a period of months from a number of vehicle types. As referenced in this GAO report, billions of dollars are spent and military readiness is degraded due to unplanned maintenance associated with US  weapons (``weapons'' includes vehicles). This degradation of resources and readiness may be significantly reduced through the use of effective predictive maintenance, as dictated in the 2002 DOD predictive maintenance policy.

\subsection{Fault Signals and US Army Vehicles}

Vehicle fault targets in this analysis are recorded in accordance with the Society of Automotive Engineers (SAE) J1939 and J1708 data specifications \cite{rit_report}. It is important to note that the analysis we propose relies on these \emph{fault signals}, also known as Diagnostic Trouble Codes (DTC), as a target and not an observation of vehicle failure. This is because failure data is not available in the analyzed set. The assumption that vehicle fault signals or DTC data is highly correlated with actual failure is supported by the detailed work regarding the CBM dataset completed by \citeN{rit_report} as described directly below:

\blockquote{DTCs can be very useful for development and enhancement of PHM (Predictive Health Monitoring) capabilities, because they contain the ground truth information for impending failures. In fact, they represent a level of condition monitoring that is already implemented by electronic control units (ECUs), employing traditionally conservative thresholds to reduce false alarm problems.}

While true vehicle failure would be an informative additional covariate, the goal of predictive maintenance is to take action based on vehicle signals before failure, where failures have high correlation with fault signals, as these signals measure the true status of critical vehicle systems (i.e. engine, transmission, and brakes).

\section{Generative Fault Prediction Model Definition}
\label{sec:cond_model}

In this section we propose a target distribution of vehicle faults conditioned on vehicle sensor values, which would provide significant utility towards predictive maintenance. We formulate a model for this target distribution employing generative learning. We fully specify this generative learning model and provide details concerning training and experimentation. The proposed model is designed to predict if a vehicle will experience a fault and time to first fault, given a fault is predicted to occur. 

\subsection{Target Distribution Definition}
\label{sec:dist_def}

The ultimate goal of this effort is to define the most accurate probability distribution, shown in Equation~\eqref{eq:prob_deepar}, where $(z_{t}, t\geq 0)$ is the fault \emph{target}, or fault indicator time series, and $(x_{i, t}, 1\leq i\leq 38, t \geq 0)$ are the sensor \emph{features} or covariate time series. Here $t$ is defined as a time point and $i$ is the label of sensor feature covariates, where there are $38$ total sensor covariates in the CBM dataset considered. With this target distribution, a vehicle operator or maintainer would know precisely the probability of \emph{if} a vehicle is to experience a fault from time point $t_0$ to $t_{0} + T$, or $t = [t_{0}:t_{0} + T
]$, and \emph{when} the fault would occur. We propose this information will allow for vehicle operators and maintainers to significantly improve predictive maintenance. Below we define this conditional probability distribution, which we name the target distribution:

\begin{equation}
\label{eq:prob_deepar}
 P(z^{k}_{t_{0}:t_{0} + T} \vert z^{k}_{0:t_{0} - 1}, (x^{k}_{i, 0:t_{0} - 1}, 1\leq i \leq 38)) \: \: \forall k \in K \: \: \forall \: t_{0} \in S, \end{equation}
where, $z^{k}_{t_{0}:t_{0} + T}$ represents the time series of fault signals collected from vehicle $k$ from time point $t_0$ to $t_0 + T$, $x^{k}_{i, t_{0}:t_{0} -1}$ represents the time series of sensor feature $i$ collected from vehicle $k$ from time point $0$ to $t_0 - 1$, and for model training and experimental purposes, the set $S$ comprises randomly selected start times for each vehicle $k$ in $K$, where $K$ is the set of selected vehicles for analysis. 

 \subsection{Generative Model Definition}
 \label{sec:model_def}
 
 While ideally we could formulate the probability distribution in Equation \eqref{eq:prob_deepar} explicitly, in practice this is not feasible so we employ a generative model to forecast outcomes given the data collected, and subsequently employ these results to model the target distribution. To accomplish this, we employ a hybrid of generative models, using DeepAR in conjunction with STAM (see Section \ref{sec:ts_pred}) and a VAE. The complete prediction process is described below:

\begin{itemize}
    \item \textbf{Step 1:} We train a DeepAR model $(Q_{\theta})$ on a large (relative to forecast length) block of vehicle sensor feature and fault target data.
    \item \textbf{Step 2:} Given time point $t_0$ to begin our forecasting period, we train the STAM model $(g_{\theta})$ and a VAE $(v_{\theta})$ to generate estimates of future sensor feature covariates.
    \item \textbf{Step 3:} We generate future targets from times $t_0$ to $t_0 + T$ using the previously trained DeepAR model ($Q_{\theta}$), and covariates generated from both the STAM model $(g_{\theta})$ and the VAE model $(v_{\theta})$. 
    \item \textbf{Step 4:} We use classifier ($C$ - see Equation \eqref{eq:class}) and regression model ($R$ - see Equation \eqref{eq:reg}) to determine if there is a fault during the forecasted time period $[t_0:t_0 + T]$, and what the time to first fault is, given a fault is expected to occur.  
\end{itemize}

\begin{figure}[h!]
    \centering
\includegraphics[width=9cm]{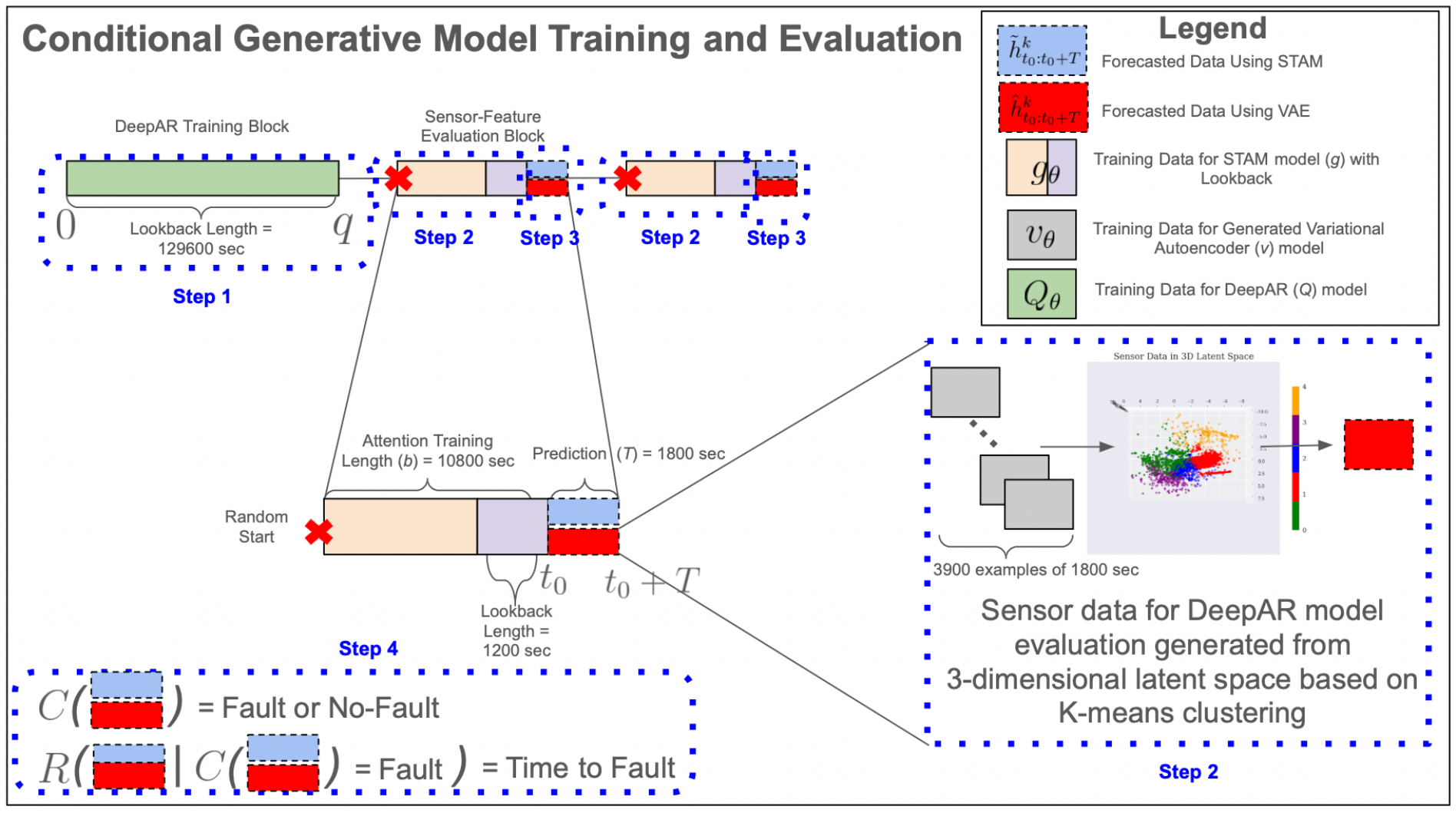}
    \caption{Overview of generative model for conditional vehicle fault prediction on vehicle $k \in K$. At Step 1, a DeepAR model ($Q_\theta$) is trained using time series of fault target and sensor feature values from time point $0$ to $q$, and is subsequently evaluated at randomly selected time periods, with $t_0 \in S$. At Step 2, a STAM model ($g_{\theta}$) is trained over a period of $b=$ 3 hours and a VAE model ($v_\theta$) is trained on out of sample data. At Step 3, the DeepAR model ($Q_\theta$) generates representations of the fault indicator $\tilde h^k_{t_0:t_0+T}$ (resp. $\hat h^k_{t_0:t_0+T}$) of $T =$ 30 minutes using the sensor feature covariates generated from the STAM model (resp. the VAE model), where STAM utilizes a lookback length of 20 minutes. At Step 4, the classifier ($C$) and the regression model ($R$) use the predictions from both the STAM and VAE model to forecast if there is a fault and what time the first fault is (given a fault is predicted) from time point $t_0$ to $t_0 + T$.}
    \label{fig:enter-label}
\end{figure}

The complete process of model training and experimentation on a single vehicle dataset is shown in Figure~\ref{fig:enter-label}, with each step identified. We note that each vehicle $k \in K$ requires the training of a DeepAR model, $Q_{\theta}$, and each forecast $t_{0} \in S$ requires the training of a STAM model, $g_{\theta}$, potentially with the evaluation of the VAE model, $v_{\theta}$. We will further explain this process in the proceeding subsections. 

\subsection{Spatio-Temporal Attention Mechanism (STAM) and Variational Auto Encoder (VAE) for Feature Generation}

As discussed in Step 1, in order to develop Equation \eqref{eq:prob_deepar}, we reference the DeepAR model, $Q_{\theta}$ and define a lookback parameter of length $\ell$ and Spatio-Temporal Attention Mechanism (STAM) training length $b$. It is critical to note that to achieve reasonable performance with the DeepAR model, \emph{features} must be available during the prediction time period $t = [t_{0}:t_{0} + T]$, where $t_{0}$ is the prediction start time ($t_0 \in S$ for each vehicle $k$ in $K$) and $t_{0} + T$ the end time for the selected forecast. The requirement of features during the prediction time poses a problem during practical forecasting because this information is \emph{future sensor} data which is not available at prediction time. To meet this requirement, we produce sensor feature data during the prediction time period using two generative methods:

\begin{enumerate}
    \item $g_{\theta} \equiv \textit{spatio-temporal attention mechanism (STAM)}$, where the STAM is trained on recent data: $g_{\theta}(x^{k}_{i, t_{0} - b:t_{0} - 1}) \Rightarrow \tilde{x}^{k}_{i, t_{0}:t_{0} + T}$. This model captures the vehicle's most recent operational history. 
    \item $v_{\theta} \equiv \textit{variational auto-encoder (VAE)}$, where the VAE is trained on out of sample data: $v_{\theta}(\hat{x}^{k'}_{i, t_{0}':t_{0}' + T}) \Rightarrow \hat{x}^{k}_{i, t_{0}:t_{0} + T}$ such that $t_{0}' \in S'$ and $k' \in K'$, and $k' \notin K$. This model captures how the vehicle will be operated in the future, which is either based upon the true data, $v_{\theta}(x^{k}_{i, t_{0}:t_{0} + T})$, or conditionally selected based on latent space cluster state, $v_{\theta}(\hat{x}^{k'}_{i, t_{0}':t_{0}' + T})$, to \emph{simulate} actual vehicle operating conditions. 
\end{enumerate}

In the prediction process, with the generated sensor data from both the STAM and VAE generators, we propose calculating two \emph{representations} of the fault indicator future time series:
\begin{subequations}
\begin{equation}
Q_{\theta}(z^{k}_{t_{0}:t_{0} + T} \vert z^{k}_{0:q}, x^{k}_{i, 0:q}, \tilde{x}^{k}_{i, t_{0}:t_{0} + T}) \Rightarrow  \tilde{h}^{k}_{t_{0}:t_{0} + T}
\label{eq:q_attn}
\end{equation}
\begin{equation}
Q_{\theta}(z^{k}_{t_{0}:t_{0} + T} \vert z^{k}_{0:q}, x^{k}_{i, 0:q}, \hat{x}^{k}_{i, t_{0}:t_{0} + T}) \Rightarrow  \hat{h}^{k}_{t_{0}:t_{0} + T}
\label{eq:q_vae}
\end{equation}
\end{subequations}
where $\tilde{h}^{k}_{t_{0}:t_{0} + T}$ and $\hat{h}^{k}_{t_{0}:t_{0} + T}$, are hidden representations of the target distribution of $z^{k}_{t_{0}:t_{0} + T}$. We call these values hidden representations because they are used in the classification and regression models, $C(\tilde{h}^{k}_{t_{0}:t_{0} + T}, \hat{h}^{k}_{t_{0}:t_{0} + T})= c \in \{0, 1\}$ and $R(\tilde{h}^{k}_{t_{0}:t_{0} + T} \vert C(\tilde{h}^{k}_{t_{0}:T}, \hat{h}^{k}_{t_{0}:t_{0} + T})= 1)= f \in \mathbb{R}^{+}$, shown in Equations \eqref{eq:class} and \eqref{eq:reg}, where the regression model $R$ is not trained with the VAE generated hidden representation, $\hat{h}^{k}_{t_{0}:t_{0} + T}$. Here the predicted value from the regression, $f$, is defined as the time to \emph{first} fault observed over the forecasted time period. We focus on predicting the time to first fault, as it has the most practical implications with respect to vehicle condition and function: the first fault is most likely to cause a maintenance event affecting operation. The classification and regression models take in the hidden representation data given each type of input and predicts fault or no-fault, and time to first fault. We employ the random forest classification and random forest regression models for $C$ and $R$, respectively. These models were selected based upon their performance and potential for interpretability \cite{random_forest}.  Additionally, the parameters of the DeepAR model,  $\theta$, are identical for the evaluation of $Q$ when used with generated covariates $\tilde{x}^{k}_{i, t_{0}:t_{0} + T}$ and $\hat{x}^{k}_{i, t_{0}:t_{0} + T}$.

\begin{equation}
C(\tilde{h}^{k}_{t_{0}:t_{0} + T}, \hat{h}^{k}_{t_{0}:t_{0} + T})= c \in \{0, 1\}.
\label{eq:class}
\end{equation}

\begin{equation}
R(\tilde{h}^{k}_{t_{0}:t_{0} + T} \vert C(\tilde{h}^{k}_{t_{0}:t_{0} + T}, \hat{h}^{k}_{t_{0}:t_{0} + T})= 1)= f \in \mathbb{R}^{+}.
\label{eq:reg}
\end{equation}

We note that when training the classification model $C$, $v_{\theta}$ is given the \emph{true} future covariates for encoding and decoding, using the trained VAE: $v_{\theta}(x^{k}_{i, t_{0}:t_{0} + T}) \Rightarrow \hat{x}^{k}_{i, t_{0}:t_{0} + T}$. This is to ensure an accurate fault classification is given when conditioning on selected latent cluster states. While in practice the \emph{true} future sensor covariate timer series, $x^{k}_{i, t_{0}:t_{0} + T}$, would not be available for the forecast, the information which characterizes the vehicle latent cluster \emph{state} would be available. Using the VAE model, we would be able to generate future fault covariates from the true vehicle state, where covariates from the common states are known to be close in the latent space.

Next, in Section \ref{sec:vae_gen}, we will discuss the random selection and generation of sensor data from the trained VAE latent space, and subsequent simulation of fault probabilities conditioned on this state. Finally, we observe that with the output of Equations \eqref{eq:class} and \eqref{eq:reg}, we may produce a prediction for ${z}^{\star k}_{t_{0}:t_{0} + T}$, or a final time series forecast indicating exactly when an \emph{initial} or first fault will occur. We propose that this is critically important information for predictive maintenance.

\subsection{Simulation using Generative Modeling Conditioned on Vehicle State}
\label{sec:vae_gen}

\begin{figure}[h!]
    \centering
\includegraphics[width=7cm]{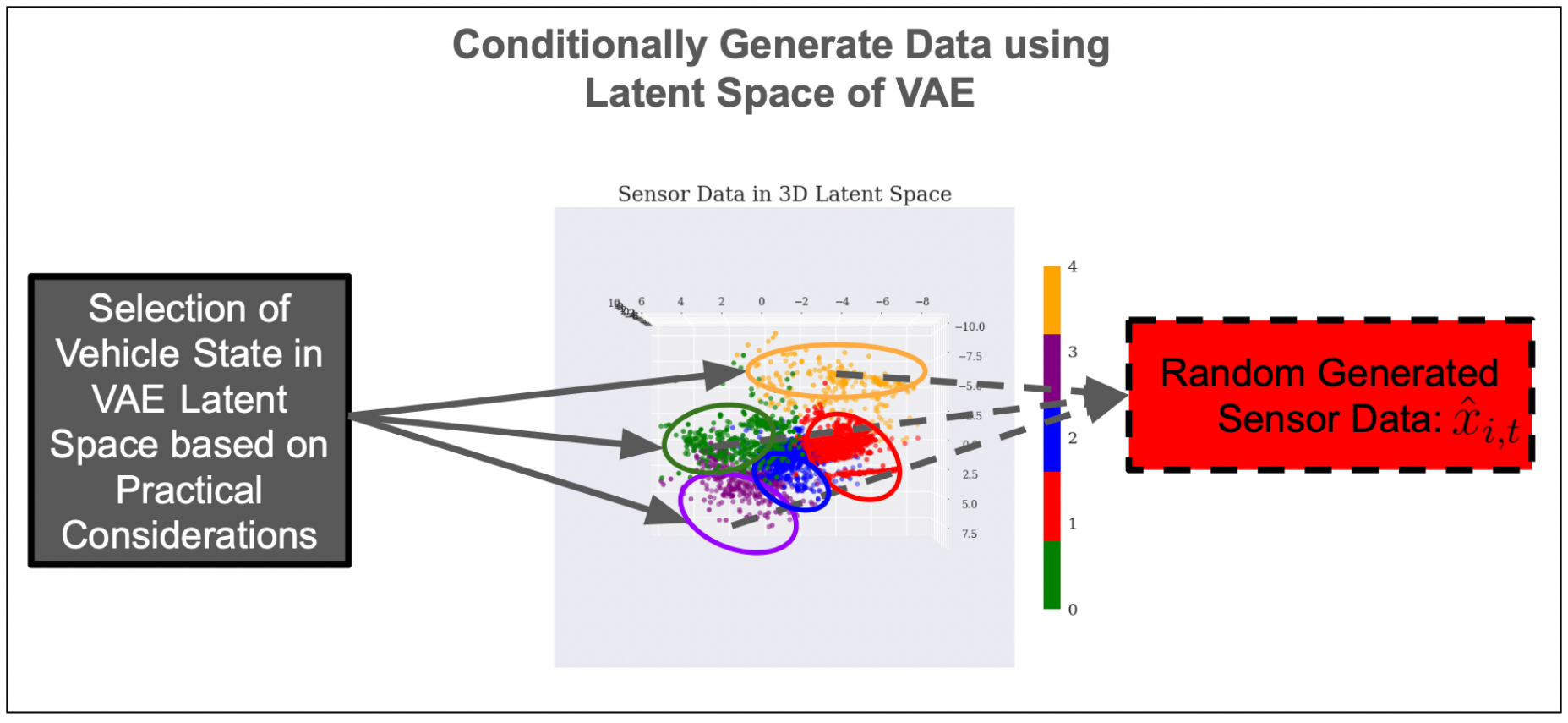}
    \caption{Illustration of mapping vehicle operational condition to generate future sensor readings.}
    \label{fig:cluster_mapping}
\end{figure}

As illustrated in Figure~\ref{fig:cluster_mapping}, the training of a VAE, $v_{\theta}$, on \emph{out of sample} vehicle data $v_{\theta}(\hat{x}^{k'}_{i, t_{0}':t_{0}' + T}) \Rightarrow \hat{x}^{k}_{i, t_{0}:t_{0} + T}$ such that $t_{0}' \in  S'$ and $k' \in K'$, and $k' \notin K$ allows us to define latent states, in a reduced dimension, which represent practical considerations associated with vehicles. These practical considerations could include: vehicle age, maintenance condition, location, etc. We enumerate the practical considerations present in in the CBM dataset and discuss their impact on the vehicle latent states in Section \ref{sec:rep_space}.  These states are designated by clustering on the reduced latent dimension produced by the encoding step of the VAE. 

In order to simulate vehicle conditions, which were not observed in the actual data, we may randomly sample data points from the selected state, decode the random point to generate sensor data, $\hat{x}^{k'}_{i, t_{0}':t_{0}' + T}$. We may then use this data in the DeepAR model and generate the hidden representation $\hat{h}^{k'}_{t_{0}':t_{0}' + T}$, with the simulated sensor values. This will produce alternate values for the classification of vehicle faults and time to first fault, using models $C$ and $R$ respectively, based on practical considerations.

\section{Results}
\label{sec:results}

We develop our results addressing two topics. First, what does the clustering of vehicle sensor data, discussed in Section \ref{sec:vae_gen}, practically indicate with respect to vehicle conditions. Second, how accurate is the discriminative model described in Section \ref{sec:dist_def} at predicting if there will be a vehicle fault observed during the generated time period, and how accurately can we predict when the first fault will occur. We provide results allowing for model interpretation, and comparison to alternate generative model formulations. We discuss the interpretation of model parameters and convergence analysis regarding the training of the DeepAR model in the appendix. 

\subsection{Simulation using Latent Space Representation and K-means Clustering}
\label{sec:rep_space}

\begin{figure}[h!]
\centering
\begin{subfigure}[b]{0.33\textwidth}
        $ $   \includegraphics[width=\textwidth]{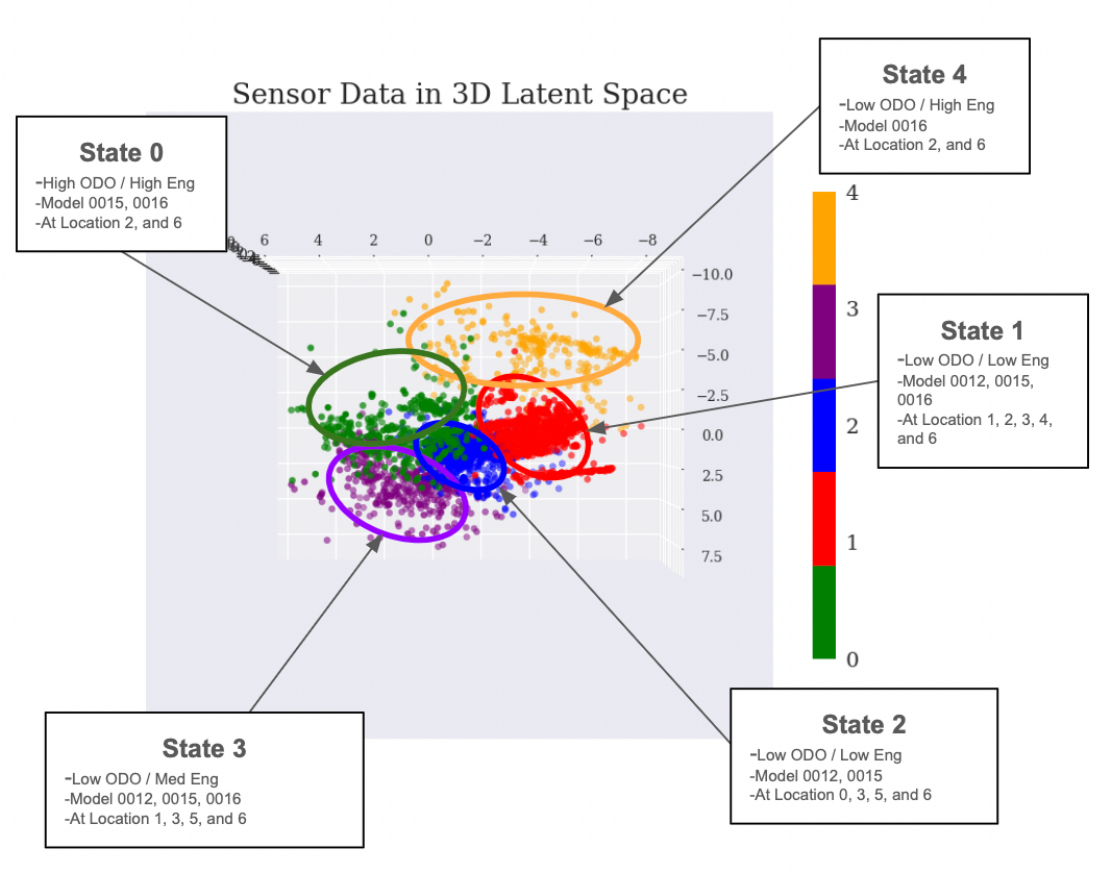}
   \caption{K-means clusters with state interpretation.}
   \label{fig:clust1} 
\end{subfigure}
\begin{subfigure}[b]{0.33\textwidth}
   \includegraphics[width=\textwidth]{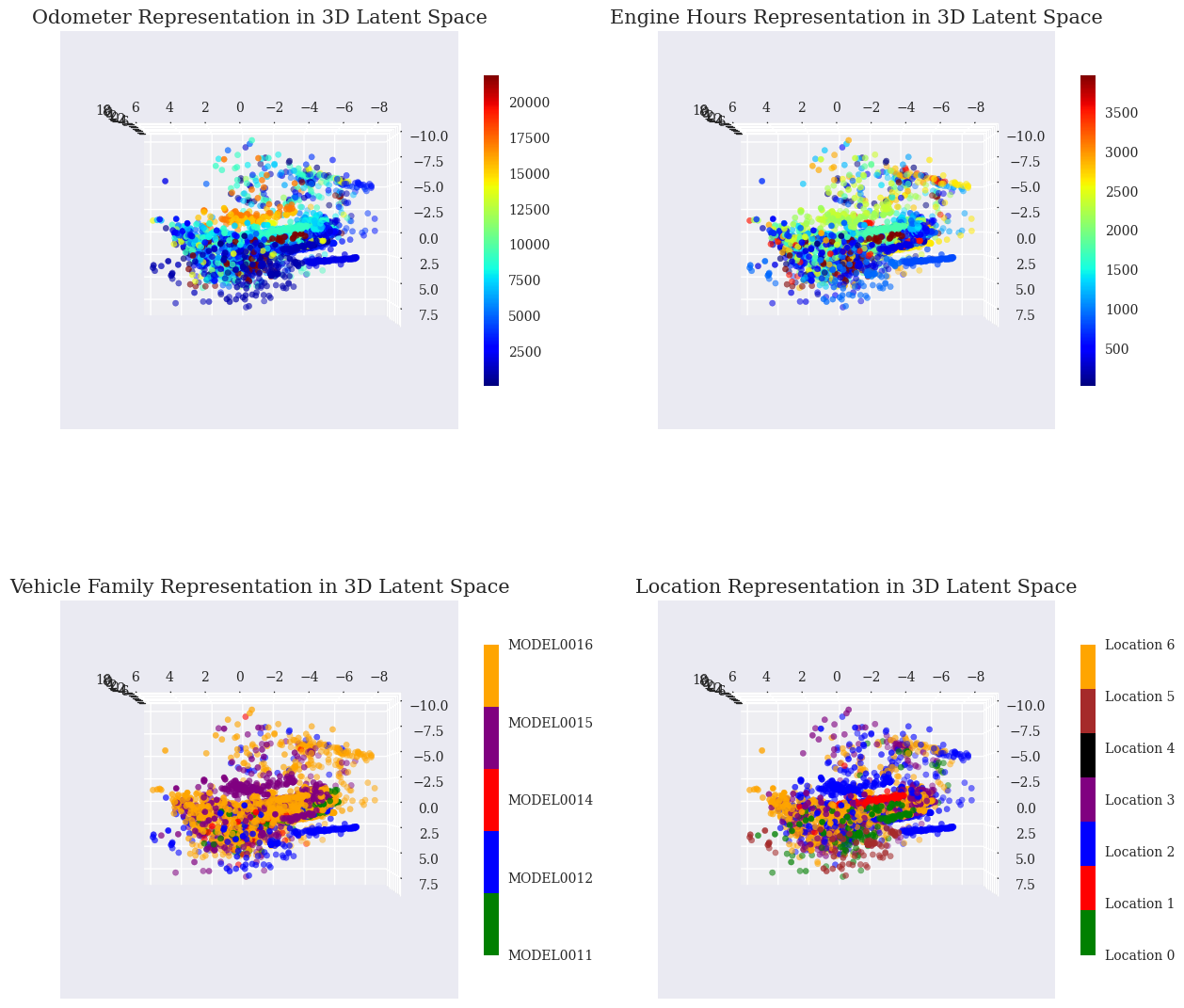}
   \caption{VAE latent representation with metadata.}
   \label{fig:clust2}
\end{subfigure}
\caption{K-means cluster representation labeled on VAE latent space with metadata interpretation.}

\label{fig:cluster_analysis}
\end{figure}

In Figure \ref{fig:cluster_analysis} we provide a representation of vehicle out of sample sensor data in a three dimensional latent space. Here the trained VAE, $v_{\theta}$, maps historical sensor readings $\hat{x}^{k'}_{i, t_{0}':t_{0}' + T}$, such that $t_{0}' \in S'$ and $k' \in K'$, into the latent representation shown. See Section \ref{sec:cond_model} for definitions of these variables. Subsequently, the K-means clustering algorithm is applied to the latent representation, producing five clusters. 

\begin{figure}
    \centering
 \includegraphics[width=0.3\linewidth]{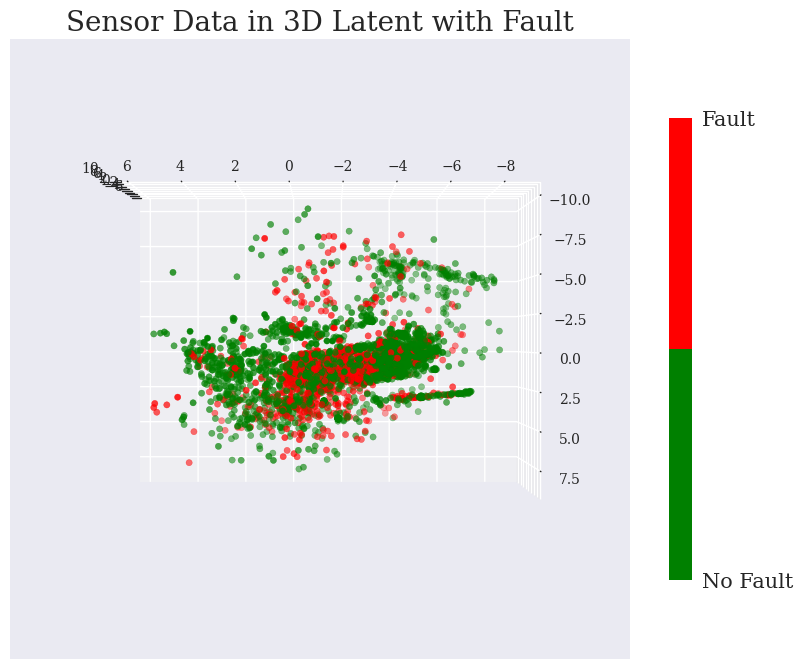}
    \captionof{figure}{Latent space fault visualization.}
\label{fig:fail_clust}
\end{figure}

We referenced metadata with respect to vehicle location, odometer mileage, engine hours, and sub-family and found that the VAE latent state mapping and K-means application identified practical clusters or \emph{states}, numbered 0-4, in which each vehicle operates. Figure \ref{fig:clust2} provides the latent representation of each of the metadata categories (location, vehicle odometer mileage, engine hours, and vehicle sub-family). This figure shows how the logical clusters or states were interpreted in Figure \ref{fig:clust1}. In Figure \ref{fig:fail_clust} we provide another view of the three dimensional latent space, where each point is labeled according to whether the vehicle truly experienced a fault, or did not experience a fault, during the generated time period $t = [t_{0}:t_{0} + T]$.

As proposed in Section \ref{sec:cond_model}, we complete a simulation generating sensor data using the decoding function of the VAE to condition the trained DeepAR model $Q_{\theta}$ when producing $\hat{h}^{k}_{t_{0}:t_{0} + T}$, the VAE generated hidden representation of the true fault indicator function $z^{k}_{t_{0}:t_{0} + T}$. We employ the trained classifier, $C$, proposed in Equation \eqref{eq:class} to determine if a vehicle experiences a fault during a predicted time period. This allows practitioners to generate simulated outcomes conditioned on practical considerations, or ``what if'' scenarios associated with their vehicles: i.e. what would happen if they changed the location the vehicle operates out of, what happens as the vehicle mileage, or engine age increase, etc. In Table \ref{fig:fail_table} we provide these simulated probabilities. In this simulation, for the same test vehicles and randomly selected time periods used throughout this analysis, we have generated sensor covariates, $\hat{x}^{k}_{t_{0}:t_{0}+T}$, by randomly sampling data points in the VAE latent space from a fixed state, where the members of the state are determined by the K-means model. For example, if all the vehicles ($k \in K$) are assumed to be operating in State 0, we may sample hundreds of points in the latent space only from this fixed state, generate $\hat{x}^{k}_{t_{0}:t_{0}+T}$ using the trained VAE $v_{\theta}$, and subsequently use this generated data in the DeepAR model $Q_{\theta}$  to produce the hidden representation $\hat{h}^{k}_{t_{0}:t_{0} + T}$, which is then evaluated using the trained classifier $C$. In this example for State 0, we expect a fault rate of $0.080$. 

\begin{table}[h!]
  \captionof{table}{Simulation of fault rates using sensor data generated from vehicle states.}
    \centering
\begin{tabular}{|c|c|}
\hline
                           \textbf{State} & \textbf{Conditional Fault Rate} \\ \hline
State 0       & 0.080            \\ \hline
State 1  & 0.078            \\ \hline
State 2 & 0.069             \\ \hline
State 3  & 0.111             \\ \hline
State 4 & 0.098             \\ \hline
\end{tabular}
\label{fig:fail_table}
\end{table}

The conditioned fault rates observed in this simulation follow logically given the practical interpretation of vehicle states given in Figure \ref{fig:clust1}. We observe, relative to other states, that the fault rate of States 3 and 4 are the highest, where these are vehicles with medium to high engine hours and lower odometer readings. This may be an indicator of scheduled service requirements for newer vehicles. Additionally, considerations with locations $1$ and $3$ may weigh on the fault rates as different locations often have varying maintenance programs. As noted in the introduction and formulation of this generative learning model, a great advantage is given by the ability to visualize and interpret results in order to understand why the model is generating specific fault forecasts. Furthermore, the values provided in Table \ref{fig:fail_table} are validated by US Army published standards and observed operational readiness rates of vehicles, cited to be between $80\% - 100\%$ \cite{orcen_report}. 

\subsection{Fault Prediction}
\label{sec:fail}

In this section, we provide a visual representation of the performance of the discriminative classifier proposed in Equation \eqref{eq:class}, which determines if a fault is going to be forecasted during the generated time period $t = [t_{0}:t_{0} + T]$. In the models trained and experiments visualized, the time period of prediction is 30 minutes, where $T = 1800 \: secs$. To be specific, we consider two cases: 1. using only the STAM generated covariate data, and 2. both the STAM and the VAE generated covariate data. In the latter case, for the VAE we use the actual latent state of the vehicle, by referencing the true future feature covariates. Using the actual latent state facilitates the development of the simulation by mapping the correct latent state to the hidden representation. The latent state is available in practice as the practitioner knows current attributes of the vehicle (vehicle mileage, engine age, etc) and may choose the appropriate state. 

\begin{figure}[htb] 
\centering
\begin{subfigure}[b]{0.3\textwidth}
        $ $   \includegraphics[width=\textwidth]{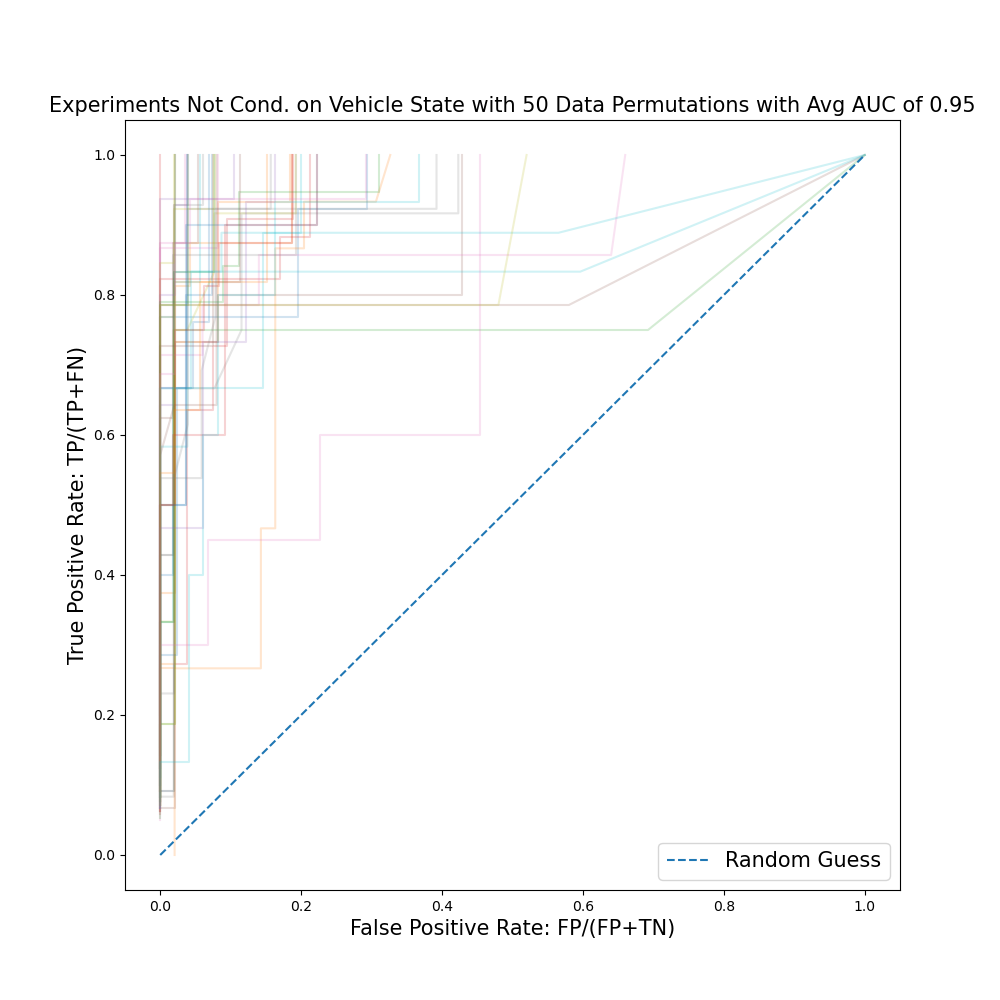}
   \caption{ROC curves using STAM covariates.}
   \label{fig:results1} 
\end{subfigure}
\begin{subfigure}[b]{0.3\textwidth}
\includegraphics[width=\textwidth]{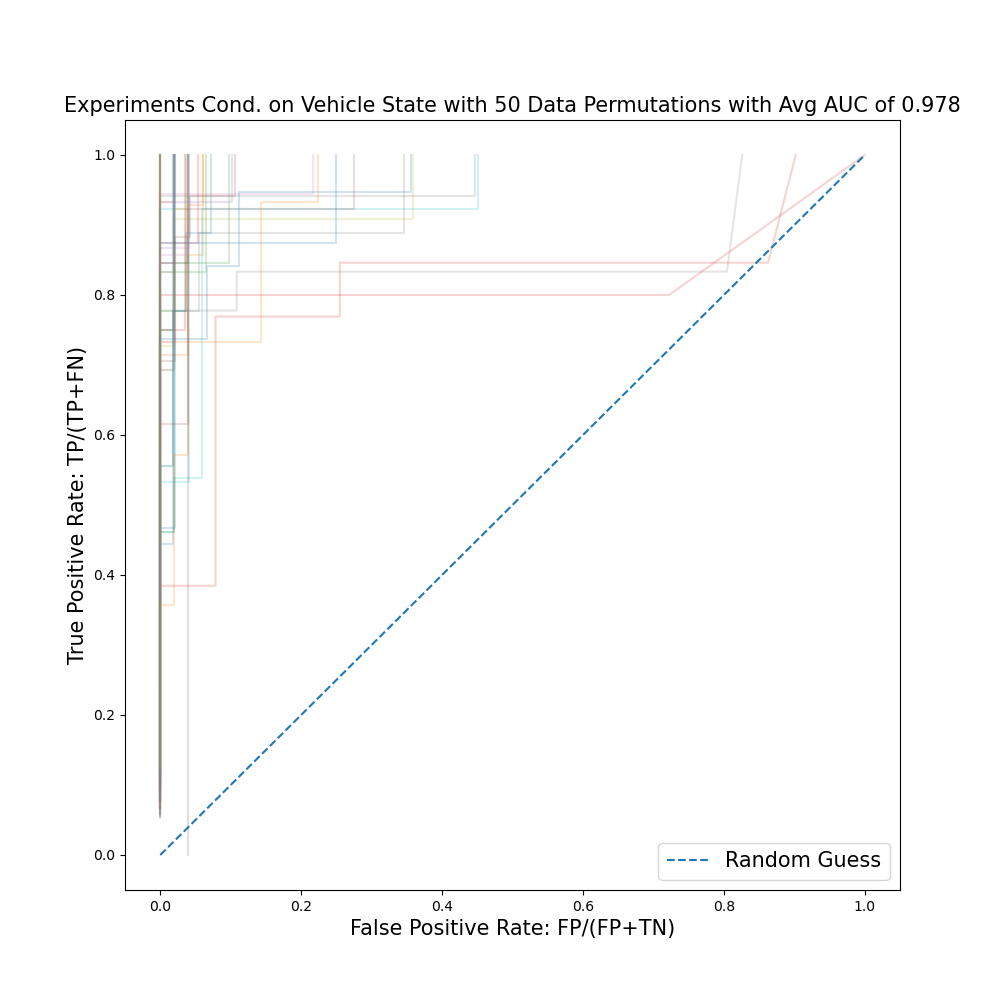}
\caption{ROC curves using STAM and VAE covariates.}
\label{fig:results2}
\end{subfigure}
\caption{ROC curves of our proposed discriminative classifiers. We choose 50 pairs of training and test data by random permutation. We plot the ROC curve on each test data using different colors after training our classifiers on the corresponding training data. We compute the average AUC for the 50 ROC curves. In Figure 6a, we use the representations of the fault indicator generated by the DeepAR model that takes in only the STAM generated sensor covariates, while in Figure 6b, we utilize the representations of the fault indicator generated by the DeepAR model that takes in both the STAM and VAE generated sensor covariates.}
\label{fig:auc}
\end{figure}

For each case, we choose 50 pairs of training and test sets through random permutation of the data. We note that the training and test sets were not split between distinct vehicles, because in practice the regression model will be updated as the vehicle is operated. As shown in Figure~\ref{fig:auc}, for the test set in each of the 50 pairs, we plot the Receiver Operating Characteristics (ROC) curve using a unique color.  The average Area Under the Curve (AUC) for the 50 tests are also computed.

These results articulate the highly accurate and steady performance achieved when using: 1. only the STAM generated covariate data, resulting in a $0.950$ AUC, and 2. both the STAM and the VAE generated covariate data, resulting in a $0.978$ AUC. More specifically, Figure \ref{fig:results1} displays the ROC curves when the hidden representation ($\tilde{h}^{k}_{t_{0}:t_{0} + T}$) is generated by the DeepAR model taking in only the STAM generated covariates (see Equation \eqref{eq:q_attn}) and this hidden representation is used in the training and testing of the classifier: $C (\tilde{h}^{k}_{t_{0}:t_{0} + T})$. In Figure \ref{fig:results2}, the hidden representations ($\tilde{h}^{k}_{t_{0}:t_{0} + T}$ and $\hat{h}^{k}_{t_{0}:t_{0} + T}$) generated by the DeepAR models using the covariates generated from the STAM and the VAE (see both Equations \eqref{eq:q_attn} and \eqref{eq:q_vae}) are both used in the training and testing of the classifier: $C (\tilde{h}^{k}_{t_{0}:t_{0} + T}, \hat{h}^{k}_{t_{0}:t_{0} + T})$. We observe that the inclusion of both the STAM and VAE generated covariates improves the average AUC performance by $2.8\%$. This improvement is expected as additional information is provided to the DeepAR ($Q_{\theta}$) model via the VAE ($v_{\theta}$), with the actual latent state information. Even without the latent state information, the model performs with satisfactory accuracy ($0.950$). To further illustrate the advantages of our method in fault prediction, we compare the AUC performance of our method to that of other existing methods in Section~\ref{sec:compare}.

\subsection{Time to First Fault Prediction}
\label{sec:ttf}

As discussed in Section \ref{sec:model_def}, once a fault is determined to occur during the predicted time period, $C(\tilde{h}^{k}_{t_{0}:t_{0} + T}, \hat{h}^{k}_{t_{0}:t_{0} + T})= 1$, we employ a random forest regression model, $R(\tilde{h}^{k}_{t_{0}:t_{0} + T} \vert C(\tilde{h}^{k}_{t_{0}:t_{0} + T}, \hat{h}^{k}_{t_{0}:t_{0} + T})= 1)= f \in \mathbb{R}^{+}$, to predict the time to first fault \cite{random_forest}. The regression model $R$ achieves a coefficient of determination ($r^2$) of $0.77$  in predicting time to first fault, where $R$ was trained on the 50th quantile generated results. We define the coefficient of determination as $r^2 = 1 - \frac{\sum_{i} (y_i - f_i)^2}{\sum_{i} (y_i - \bar{y})^2}$, where $y_i$ is the true time to first fault, $f_i$ is the forecasted time to first fault for observation $i$, and $\bar{y}$ is the true mean. We note that all true time to first fault values are no further than 4 minutes ($240\: secs$) from the forecasted time to first fault; however, these predictions could benefit from further refinement on robustness since the present prediction indicates potential under-estimation of the time to first fault. This issue has practical implications for predictive maintenance and we seek to address this shortcoming in future work.

\subsection{Comparison to Alternate Models on Fault Prediction}
\label{sec:compare}

We selected two methods for comparison to the the proposed model. The selected methods include a LSTM and a STAM model, where both are used to directly generate hidden representations of the target fault indicator. By contrast, our proposed method uses the STAM model to generate sensor covariates, and then inputs the generated sensor covariates into the DeepAR model to generate hidden representations of the target fault indicator.

\begin{figure}[htb]
\centering
\begin{subfigure}[b]{0.3\textwidth}
        $ $   \includegraphics[width=\textwidth]{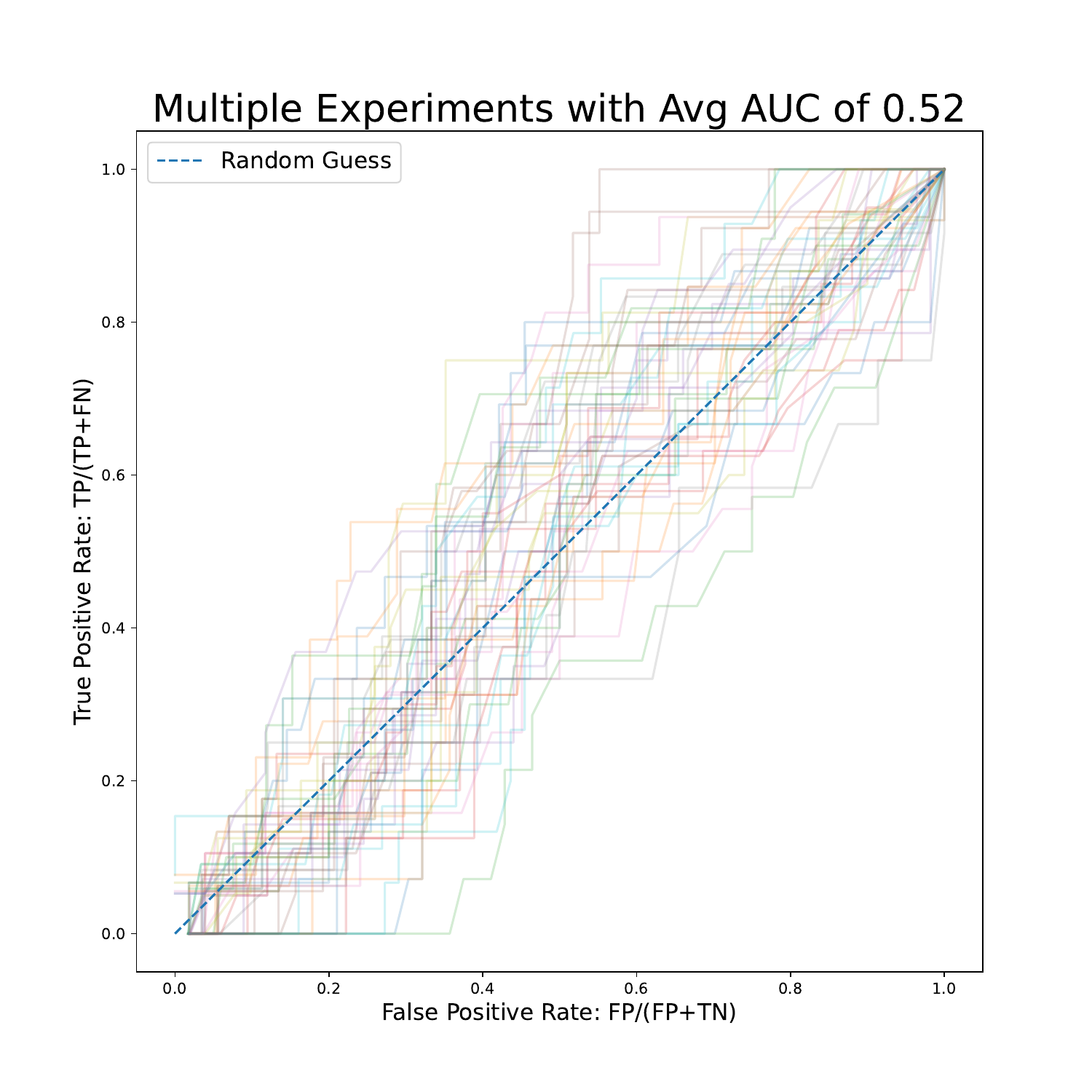}
   \caption{ROC curves of a LSTM-based classifier.}
   \label{fig:results_lstm} 
\end{subfigure}
\begin{subfigure}[b]{0.3\textwidth}
   \includegraphics[width=\textwidth]{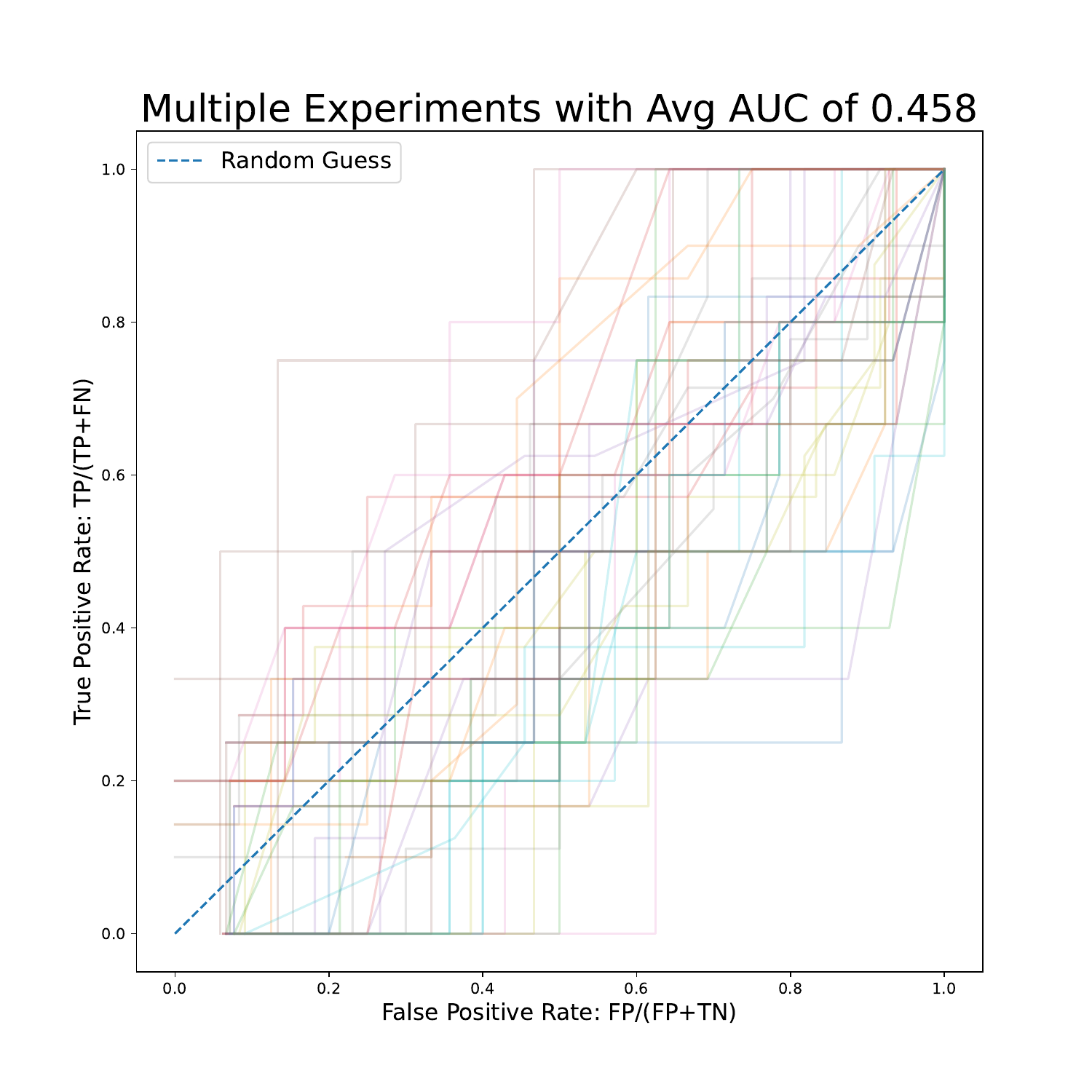}
   \caption{ROC curves of a STAM-based classifier.}
   \label{fig:results_sparse}
\end{subfigure}

\caption{ROC curves of alternate classification models. We choose 50 pairs of training and test data by random permutation. We plot the ROC curve on each test data using different colors after training our classifiers on the corresponding training data. We compute the average AUC for the 50 ROC curves. In Figure 7a, we we utilize the representations of the fault indicator directly generated by LSTM, while in Figure 7b, we we utilize the representations of the fault indicator directly generated by STAM.}
\label{fig:model_compare_figs}
\end{figure}

The time period is set to be $t = [t_0:t_0 + T]$. As with the proposed methodology, the directly generated hidden representation, denoted as $\tilde{h}^{k}_{t_0:t_0 + T}$, serves as the input into a classification model, $C(\tilde{h}^{k}_{t_0:t_0 + T}) = c \in \{ 0, 1\}$. The performance of the comparison models are visualized through plots of ROC curves in Figure~\ref{fig:model_compare_figs}. In Table~\ref{fig:compare_table}, we compare the average AUC for these methods and see clearly that the performance for both the LSTM and STAM models is significantly poorer when compared to the proposed model.

\begin{table}[h!]
\caption{Model comparison table.}
\centering
\begin{tblr}{|c|c|}
\hline
\textbf{Model for Comparison} & \textbf{AUC} \\ \hline
LSTM Baseline & 0.520           \\ \hline
STAM Baseline & 0.458            \\ \hline
\SetCell{h,4.5cm} DeepAR with STAM Covariates (ours)          & 0.950             \\ \hline
\SetCell{h,4.5cm} \textbf{DeepAR with STAM and VAE Covariates (ours)}    & \textbf{0.978}             \\ \hline
\end{tblr}
\label{fig:compare_table}
\end{table}

\subsection{Discussion}

A clear question arises from the proposed model: why does the use of the STAM method to generate covariates, and subsequently generating target values using the DeepAR model produce such an accurate fault representation? This is clearly true when compared to other baseline methods as presented in Figure \ref{fig:model_compare_figs}. We believe this is due to several attributes of both the DeepAR and STAM models, namely: (i) DeepAR, being auto-regressive and recurrent, is particularly designed to learn seasonal behavior well, which is specifically helpful with larger and longer time series datasets (ii) DeepAR incorporates the use of covariates focused towards learning a specific target distribution, (iii) STAM is demonstrated to be particularly scalable and accurate when forecasting time series values with multiple variables. 

Specifically referencing point (i), we know that the vehicle sensor feature and fault target data exhibits seasonality or cyclic behavior when explored over significant periods of time. As such, we trained the DeepAR over 36 hours of data (129600 data points). This allowed for the auto-regressive network to learn sufficient detail of recurrent structure over the fault targets and sensor feature covariates, as specified in point (ii). Additionally, while alone not accurate when predicting target values as shown in Section \ref{sec:compare}, the STAM model is excellent at generating the covariates when trained over modest time periods, such as 3 hours of data in the proposed model (10800 data points). STAM provides sufficient accuracy when forecasting the subsequent 30 minutes (1800 data points) of sensor feature data to produce fault target data which is sufficiently distinguishable to support highly accurate binary classification and regression results. Attributes of the STAM and DeepAR models are combined with appropriate training periods to leverage the strengths of both methods in order to facilitate forecasts that are well suited for this vehicle fault prediction problem.

\section{Conclusion and Further Work}

We proposed a modeling approach which allows for the highly accurate prediction of future vehicle faults, and the simulation of future faults conditioned on practical considerations. The simulated results are validated using logical interpretation of vehicle metadata and published documentation. We demonstrated that the model is interpretable and provides actionable results for practitioners in the vehicle operation and maintenance fields. Significant further work remains including:

\begin{enumerate}
    \item Improving robustness of time to first fault forecast (Section~\ref{sec:ttf}) with mitigation of underestimation.
    \item Extending prediction time to the scale of hours.
    \item Simulating the evolution of states for multiple vehicles to streamline fleet maintenance scheduling.
\end{enumerate}

We note that a link to a github repository with all modeling code and processor information is available in Appendix \ref{app:code}. The work presented represents an initial step in this modeling effort, and we will continue to update our procedures to produce accurate and interpretable results for the predictive maintenance effort. 

\section{Acknowledgement}

The material in this paper is based upon work supported by the Air Force Office of Scientific Research under award number FA9550-20-1-0397.

\newpage
\appendix

\section{Model Interpretability: Attention Weight Analysis}
\label{app:attn_wt_sect}

\begin{figure}[!htbp]
    \centering
\includegraphics[width=15cm]{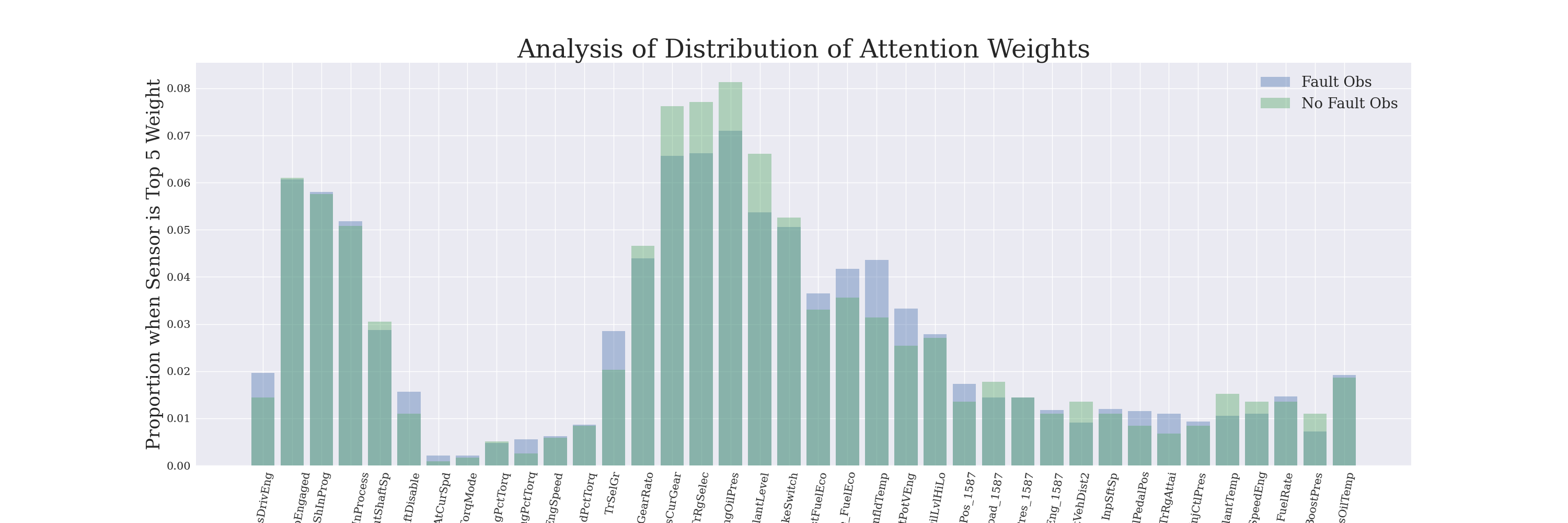}
\caption{Attention weight analysis}
\label{fig:attn_wt}
\end{figure}

An important attribute of the proposed analysis, given the methods used, is the models provide interpretable parameters. These parameters are produced during the training of the STAM neural network, used to generate future sensor feature covariates \shortcite{attn_mech}. The STAM produces attention weights when training, where these weights may be interpreted as an importance metric for each of the covariates or sensors, when generating future values. In Figure \ref{fig:attn_wt}, we visualize a mean of recorded attention weights, averaged when executing the experiments presented in Sections \ref{sec:fail} and \ref{sec:ttf}, over the randomly sampled time periods for all $k \in K$ and $t_0 \in S$. Two sets of attention weights are visualized: one set over the samples where a fault is observed during the time period, and another where a fault is not observed. 

We observe a shift in attention weights when there is a fault observed. Specifically, the sensors associated with the transmission subsystem become less important and the engine subsystem becomes more important. These results follow logically, as the engine subsystem is a more complicated component of the vehicle with numerous components and potential points of fault. These results are shown in Table \ref{fig:attn_wt_table}.

\begin{table}[h!]
\caption{Attention weight comparison table}
\centering
\begin{tabular}{|c|c|}
\hline
                           & Attention Weight \\ \hline
Engine Without Fault       & 0.480           \\ \hline
Transmission Without Fault & 0.520            \\ \hline
Engine With Fault          & 0.515             \\ \hline
Transmission With Fault    & 0.485             \\ \hline
\end{tabular}
\label{fig:attn_wt_table}
\end{table}

\section{Implementation Code}
\label{app:code}

The code base, written in python, is available at the following github link:  \href{https://github.com/patrick-kuiper/gen_pm}{generative predictive maintenance github code}. There is only one vehicle dataset provided for demonstration purposes at this github repository. The full dataset was not provided due to file size considerations, and maybe referenced for reproducibility by contacting the  following email: \href{mailto:patrick.kendal.kuiper@gmail.com}{author email}. The analysis referenced in the paper was completed with 200 vehicles of data. Training of the STAM and VAE models were completed on NVIDIA GeForce RTX 3090 GPUs with CUDA Version 12.2.  Further details may be referenced in the linked github repository.

\section{Model Training Convergence Analysis}
\label{app:loss_analysis}

We provide a visualization of the loss history recorded during the training of the DeepAR model. This allows us to confirm the DeepAR training convergence visually. Given that we trained a single DeepAR model for each vehicle analyzed in the CBM dataset, we have many training loss profiles. We see clearly the convergence of the training, for all DeepAR models, over the 100 epoch training period.

\begin{figure}[!htbp]
    \centering
\includegraphics[width=15cm]{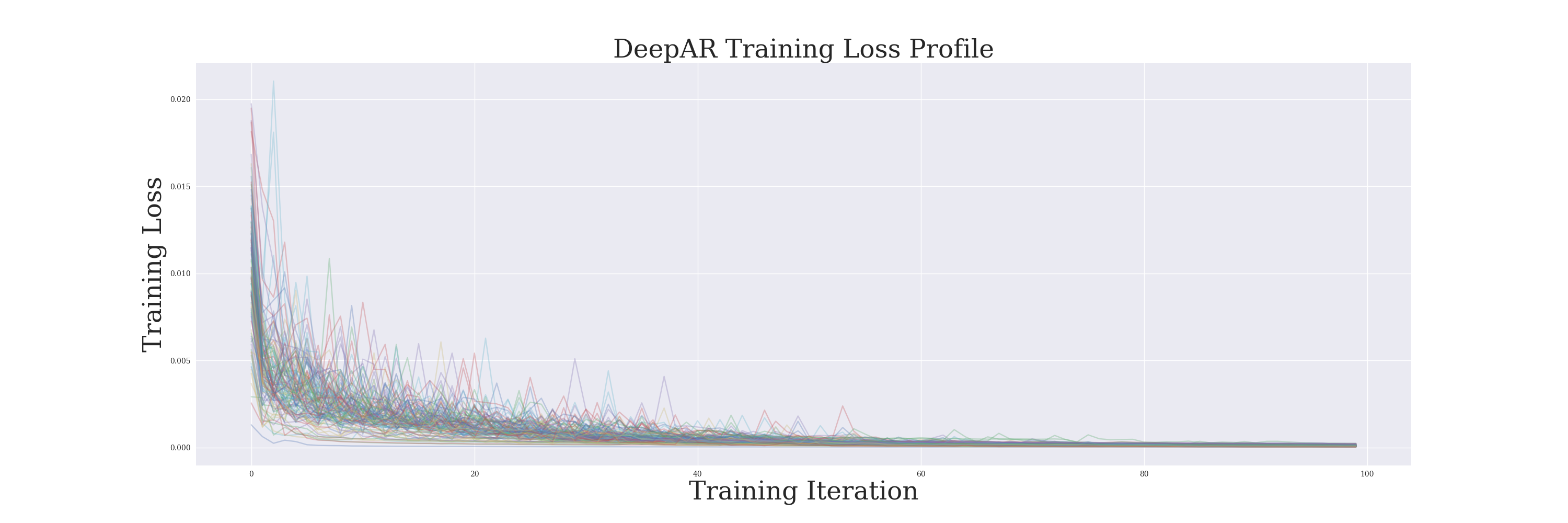}
    \caption{Training loss profile}
    \label{fig:train_hist}
\end{figure}

\footnotesize

\bibliographystyle{wsc}
\newpage
\bibliography{main_citations}

\section*{AUTHOR BIOGRAPHIES}

\noindent {\bf \MakeUppercase{Patrick Kuiper}} received his B.S. in Operations Research from The United States Military Academy at West Point in 2007, and a M.Eng. in Applied Mathematics from Harvard University in 2014. He is an active duty officer in the United States Army and served as an instructor in the West Point Department of Mathematical Sciences from 2016 to 2019. He is currently pursuing a Ph.D. in Electrical and Computer Engineering at Duke University. His research interests include applications of generative modeling and extreme value theory. His email address is \email{patrick.kuiper@duke.edu}.\\

\noindent {\bf \MakeUppercase{Sirui Lin}} is a Ph.D. student at the Department of Management Science and Engineering (MS\&E) at Stanford University, with research interest in applied probability and optimal transport-based statistical inference. His email address is \email{siruilin@stanford.edu}.\\

\noindent {\bf \MakeUppercase{Jose Blanchet}} 
is a Professor of Management Science and Engineering (MS\&E) at Stanford. Prior to joining Stanford, he was a professor at Columbia (Industrial Engineering and Operations Research, and Statistics, 2008-2017), and before that he taught at Harvard (Statistics, 2004-2008). Jose is a recipient of the 2010 Erlang Prize and several best publication awards in areas such as applied probability, simulation, operations management, and revenue management. He also received a Presidential Early Career Award for Scientists and Engineers in 2010. He is the Area Editor of Stochastic Models in Mathematics of Operations Research and has served on the editorial board of Advances in Applied Probability, Bernoulli, Extremes, Insurance: Mathematics and Economics, Journal of Applied Probability, Queueing Systems: Theory and Applications, and Stochastic Systems, among others. His email address is \email{jose.blanchet@stanford.edu}.
\\

\noindent {\bf \MakeUppercase{Vahid Tarokh}} worked at the AT\&T Labs-Research until 2000. From 2000 to 2002 he was an Associate Professor at Massachusetts Institute of Technology (MIT). In 2002 he joined Harvard University as a Hammond Vinton Hayes Senior Fellow of Electrical Engineering and Perkins Professor in Applied Mathematics. He joined Duke University in 2018 as the Rhodes Family Professor in Electrical and Computer Engineering, Computer Science, and Mathematics and Bass Connections Endowed Professor. He was also a Gordon Moore Distinguished Research Fellow at CALTECH in 2018. Since Jan 2019, he has also been named as a Microsoft Data Science
Investigator at Duke University. His email address is \email{vahid.tarokh@duke.edu}.

\end{document}

%% file: wscbib.tex
\makeatletter
\let\@internalcite\cite
\def\cite{\def\@citeseppen{-1000}%
    \def\@cite##1##2{(##1\if@tempswa , ##2\fi)}%
    \def\citeauthoryear##1##2##3{##1 ##3}\@internalcite}
\def\citeNP{\def\@citeseppen{-1000}%
    \def\@cite##1##2{##1\if@tempswa , ##2\fi}%
    \def\citeauthoryear##1##2##3{##1 ##3}\@internalcite}
\def\citeN{\def\@citeseppen{-1000}%
    \def\@cite##1##2{##1\if@tempswa, ##2)\else{}\fi}%
    \def\citeauthoryear##1##2##3{##1 (##3)}\@citedata}
\def\citeA{\def\@citeseppen{-1000}%
    \def\@cite##1##2{(##1\if@tempswa , ##2\fi)}%
    \def\citeauthoryear##1##2##3{##1}\@internalcite}
\def\citeANP{\def\@citeseppen{-1000}%
    \def\@cite##1##2{##1\if@tempswa , ##2\fi}%
    \def\citeauthoryear##1##2##3{##1}\@internalcite}
\def\shortcite{\def\@citeseppen{-1000}%
    \def\@cite##1##2{(##1\if@tempswa , ##2\fi)}%
    \def\citeauthoryear##1##2##3{##2 ##3}\@internalcite}
\def\shortciteNP{\def\@citeseppen{-1000}%
    \def\@cite##1##2{##1\if@tempswa , ##2\fi}%
    \def\citeauthoryear##1##2##3{##2 ##3}\@internalcite}
\def\shortciteN{\def\@citeseppen{-1000}%
    \def\@cite##1##2{##1\if@tempswa, ##2\else{}\fi}%
    \def\citeauthoryear##1##2##3{##2 (##3)}\@citedata}
\def\shortciteA{\def\@citeseppen{-1000}%
    \def\@cite##1##2{(##1\if@tempswa , ##2\fi)}%
    \def\citeauthoryear##1##2##3{##2}\@internalcite}
\def\shortciteANP{\def\@citeseppen{-1000}%
    \def\@cite##1##2{##1\if@tempswa , ##2\fi}%
    \def\citeauthoryear##1##2##3{##2}\@internalcite}
\def\citeyear{\def\@citeseppen{-1000}%
    \def\@cite##1##2{(##1\if@tempswa , ##2\fi)}%
    \def\citeauthoryear##1##2##3{##3}\@citedata}
\def\citeyearNP{\def\@citeseppen{-1000}%
    \def\@cite##1##2{##1\if@tempswa , ##2\fi}%
    \def\citeauthoryear##1##2##3{##3}\@citedata}
%
%
%
\def\@citedata{%
    \@ifnextchar [{\@tempswatrue\@citedatax}%
                  {\@tempswafalse\@citedatax[]}%
}

\def\@citedatax[#1]#2{%
\if@filesw\immediate\write\@auxout{\string\citation{#2}}\fi%
  \def\@citea{}\@cite{\@for\@citeb:=#2\do%
    {\@citea\def\@citea{, }\@ifundefined
       {b@\@citeb}{{\bf ?}%
       \@warning{Citation `\@citeb' on page \thepage \space undefined}}%
{\csname b@\@citeb\endcsname}}}{#1}}%

%
\def\@citex[#1]#2{%
\if@filesw\immediate\write\@auxout{\string\citation{#2}}\fi%
  \def\@citea{}\@cite{\@for\@citeb:=#2\do%
    {\@citea\def\@citea{; }\@ifundefined
       {b@\@citeb}{{\bf ?}%
       \@warning{Citation `\@citeb' on page \thepage \space undefined}}%
{\csname b@\@citeb\endcsname}}}{#1}}%

%
\def\@biblabel#1{}
\makeatother



\newdimen\bibindent
\bibindent=0.0em
\def\thebibliography#1{\section*{\refname}\list
   {}{\settowidth\labelwidth{[#1]}
   \leftmargin\parindent
   \itemindent -\parindent
   \listparindent \itemindent
   \itemsep 0pt
   \parsep 0pt}
   \def\newblock{}
   \sloppy
   \sfcode`\.=1000\relax}